# A Novel Efficient Approach with Data-Adaptive Capability for OMP-based Sparse Subspace Clustering

**Jiaqiyu Zhan, Zhiqiang Bai, Yuesheng Zhu***
Communication and Information Security Lab,
Shenzhen Graduate School, Peking University, China
{zjqy0429, zqbai, zhuys}@pku.edu.cn

## Abstract

Orthogonal Matching Pursuit (OMP) plays an important role in data science and its applications such as sparse subspace clustering and image processing. However, the existing OMP-based approaches lack of data adaptiveness so that the data cannot be represented well enough and may lose the accuracy. This paper proposes a novel approach to enhance the data-adaptive capability for OMP-based sparse subspace clustering. In our method a parameter selection process is developed to adjust the parameters based on the data distribution for information representation. Our theoretical analysis indicates that the parameter selection process can efficiently coordinate with any OMP-based methods to improve the clustering performance. Also a new Self-Expressive-Affinity (SEA) ratio metric is defined to measure the sparse representation conversion efficiency for spectral clustering to obtain data segmentations. Our experiments show that proposed approach can achieve better performances compared with other OMP-based sparse subspace clustering algorithms in terms of clustering accuracy, SEA ratio and representation quality, also keep the time efficiency and anti-noise ability.

## 1  Introduction

Clustering is a common task in the fields of data science, image processing, machine learning, and computer vision, etc. In the big data applications, the analysis and processing of large-scale high-dimensional data becomes an important research direction [1-2].

In general, to represent complex data, a reasonable assumption is that high-dimensional data lie in a union of multiple low-dimensional subspaces. Sparse subspace clustering (SSC) [3] algorithms have become popular to segment the high-dimensional data from different subspaces into essentially low-dimensional subspaces and reflect the underlying nature of the data. It works by creating representation coefficients of data in low-dimensional subspace for compression to build sparse representation, construct affinity matrix based on the representation coefficient matrix, and obtain the clustering label of the data by spectral clustering [4-6].

Orthogonal matching pursuit (OMP) and its improved algorithms play an important role in data science and its applications such as sparse subspace clustering [7-12]. Different from the basic SSC algorithm which uses convex programming tools to obtain the representation matrix, the OMP-based algorithms can relax problem requirements, decompose each data on the union of all data points and orthogonalize all the selected atoms at each step during decomposition, achieve faster convergence and offer better trade-off between clustering accuracy and computational efficiency.

However, the existing OMP-based approaches lack of data adaptiveness. When OMP-based algorithms are used to solve SSC problems [13-15], the parameters are often set manually by experience or according to the evaluation results after enumerating a group of optional parameters. The parameters do not adjust accordingly to the meaningful knowledge from the underlying data and remain the same for every data point regardless of its nature. It is noted that in the OMP-based algorithms when the data lies on the boundary between different categories, its dictionary atoms may be chosen from several subspaces to represent it. Since property of the dictionary is important for sparse representation [16], in this case, the data may be mislabeled resulting in low representation quality, low clustering accuracy and waste of computation resources.

Theoretically, by adding a data-sensitive selection step, the nondata-adaptive methods could become data-adaptive [17-19]. It is the motivation for this paper to improve accuracy and preserve time efficiency of OMP-based Sparse Subspace Clustering. In this paper a novel approach to enhance the data-adaptive capability for OMP-based sparse subspace clustering is proposed, also a parameter selection process to adjust parameters based on data distribution is developed to expand the existing framework. Meanwhile, a specific parameter processing algorithm is proposed. After processing parameters, the framework pays more attention to data that meets the potential laws of its belonged category and reduce the consumption when processing data that is far away from its belonged category in the processing of high-dimensional multi-category data.

The contributions of this paper are as follows:
- We propose a new approach for OMP-based sparse subspace clustering
  which add a parameter selection process in front of the existing framework.
- We propose a parameter processing algorithm in the parameter selection process,
  dealing with the important parameter of OMP-based algorithms: dictionary size.
- We define a new metric Self-Expressive-Affinity (SEA) ratio to measure the sparse
  representation conversion efficiency for spectral clustering to obtain data segmentations.

The rest of this paper is organized as follows: Section 2 briefly introduces the previous SSC framework and proposes a new approach by adding a parameter selection process. Section 3 proposes a parameter processing algorithm, and gives relevant explanations. Section 4 introduces some widely used evaluation metrics of SSC problems, and defines the SEA ratio to measure the sparse representation conversion efficiency from self-expressive matrix to affinity matrix. Section 5 verifies our theoretical analysis through experiments on real-world datasets, effectiveness, efficiency and anti-noise ability of our proposed approach and algorithm. Finally, Section 6 concludes the paper.

## 2    Data-adaptive approach for OMP-based SSC

In this section, we briefly introduce the previous sparse subspace clustering framework and proposes a new data-adaptive approach by adding a parameter selection process in front of the existing framework, then give an example to illustrate difference between previous framework and our approach.

### 2.1    Previous framework

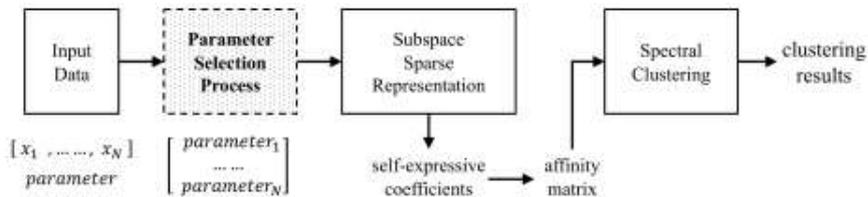

Figure 1: New approach with SSC framework

Previous basic framework of SSC can be described as: establish a subspace representation model for given data, search the representation coefficients of the data in low-dimensional subspace, then construct the affinity matrix according to the representation coefficient matrix, finally obtain the clustering results by spectral clustering method.

There are three major steps in the previous basic framework of sparse subspace clustering: data input, subspace sparse representation and spectral clustering. As shown in Figure 1, $\mathcal{X} = [x_1, \ldots, x_N]$ is the input data, $\forall x_i \in \mathcal{X}(i = 1, \ldots, N)$, a set $\mathcal{X}'_i$ is selected from $\mathcal{X}_{-i}$ due to $\mathcal{X}$ and preset parameters, where $\mathcal{X}_{-i}$ denotes $\mathcal{X}$ with the point $x_i$ removed(avoid trivial solution). If $x_i$ belongs to a set $\mathcal{S}_i$(low-dimensional subspace, denotes a category), ideally $\mathcal{X}'_i \subset \mathcal{S}_i$, then the sparse representation of $x_i$ is called subspace preserving. $c_i$ denotes representation coefficients, comes from the projection of $x_i$ onto the subspace spanned by points in $\mathcal{X}'_i$, define self-expressive coefficient matrix $\mathcal{C} = [c_1, \ldots, c_N]$, affinity matrix $\mathcal{A}$ is constructed according to $\mathcal{C}$, usually $\mathcal{A} = |\mathcal{C}| + |\mathcal{C}^T|$. Finally, applying spectral clustering [20] to A to obtain data segmentations.

Parameters in previous framework are often set manually. No matter how to set, parameters do not adjust accordingly to the meaningful knowledge from the underlying data and remain the same for every data point regardless of its nature. It is noted that in the OMP-based algorithms when the data lies on the boundary between different categories, its dictionary atoms may be chosen from several subspaces to represent it. Since property of the dictionary is important for sparse representation, in this case, the data may be mislabeled resulting in low representation quality, low clustering accuracy and waste of computation resources.

## 2.2 Our approach

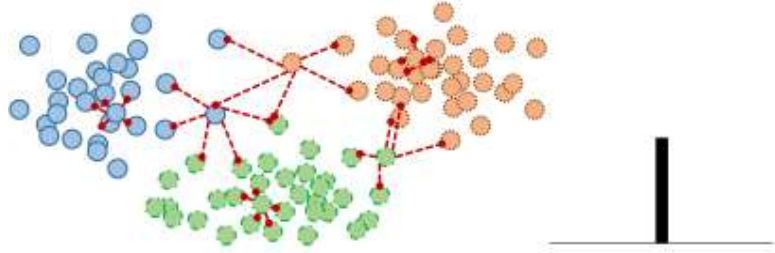

Figure 2: illustration of previous framework

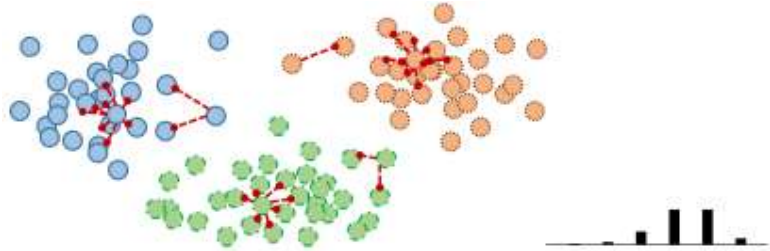

Figure 3: illustration of our approach

In our proposed new approach, we add a parameter selection process in front of existing framework for OMP-based sparse subspace clustering, after the data input step and before the subspace sparse representation step (Figure 1). In anticipation of future algorithms that fit in our approach, we suppose they do parameter selection by observing data X's attributes such as distribution, and dealing with the originally "static" parameters, make parameters more in line with the underlying nature of data.

In previous framework, all data points share same parameters; in our approach, every data point is the owner of its exclusive parameters, the exclusiveness comes from parameter selection process based on data X's underlying nature. Our approach makes the framework pays more attention to data that meets the potential laws of its belonged category.

Now we give an example to illustrate difference between previous framework and our approach. As shown in Figure 2 and Figure 3, circles in different colors and border styles represent data points belong to different clusters, red dashed line represent the selection of each point choosing atoms to build dictionary for self-expression. Histogram in Figure 2 shows dictionary size of all points, under previous framework, dictionary size of all points are the same. Thus we observe some points, e.g. lie in the fuzzy region or on the boundary between different categories, are quite possible to be mislabeled due to their dictionary atoms are chosen from several subspaces. Based on our parameter selection process, we make adaptive adjustment to dictionary size, histogram in Figure 3 shows dictionary size. When it is adjusted accordingly to the distribution of underlying data, each point choosing atoms more wisely and reduce the chances of making clustering error.

## 3 New algorithm for data-adaptive dictionary size selection

In this section, we propose a parameter processing algorithm, Algorithm 1, in the parameter selection process dealing with an important parameter of OMP-based algorithms: dictionary size.

**Algorithm 1: Parameter-K Processing**

**Input:** matrix $X = [x_1, \cdots, x_N]$, dictionary size $K$.
1: DIST $\leftarrow X^T X$.               // matrix product of $X^T$ and $X$
2: Sort all rows of DIST in descending order.
3: DIST $\leftarrow$ DIST( : , 2:K ).     // Take 2~K columns of DIST
4: DIST $\leftarrow$ mean of each row of DIST,
   MaxD $\leftarrow$ max value among DIST,
   MinD $\leftarrow$ min value among DIST.
5: Update DIST $\leftarrow$ DIST $'$ by (1).
6: Calculate $K_{array}$ by (2).
**Output:** an array of new dictionary size: $K_{array}$

The key idea of Algorithm 1 is to let parameter-K do accordingly adjustments based on data distribution to enhance the data-adaptive capability. When processing high-dimensional multi-category data, Algorithm 1 let OMP-based SSC framework pays more attention to data that meets the potential laws of its belonged category and let them choose more atoms to express themselves. And pays less attention to data that lie in the fuzzy region or on the boundary between different categories and is more likely to be mislabeled, downsize their dictionary, suppress their expression.

There are two inputs in Algorithm 1, i.e. data matrix $X \in \mathbb{R}^{dim \times N}$ and dictionary size $K \in \mathbb{N}^+$, where n denotes data dimension and N denotes the number of data points. Firstly, we get the distance between every data point and its nearest K-1 points but itself (avoid trivial solution), denoted by a matrix $DIST \in \mathbb{R}^{N \times (K-1)}$, using step 1~3 in Algorithm 1. If all data points have been normalized to same length, then by "nearest" we mean "have smallest angle between two points".

$$DIST' = K \times \frac{DIST - MinD}{MaxD - MinD} \qquad (1)$$

Secondly, we adjust DIST based on data distribution using step 4~5. After (1), DIST is normalized to [0,K], great value in DIST indicates that point is gathered densely by points of its belonged category and meets the potential laws, on the contrary, low value indicates that point is more likely to be mislabeled, so we suppress its expression.

$$K_{array} = K - round(mean(DIST)) + round(DIST) \qquad (2)$$

Finally, by (2), we add an offset, K-round(mean(DIST)), to the normalized DIST to make its mean value approximately equals to K. The *i*th element of $K_{array}$ denotes the exclusive dictionary size of $x_i$. (2) guarantees that every dictionary size is an integer, and the smallest one is positive.

Algorithm 2 presents the procedure of SSC-OMP which indicates the previous framework. Algorithm 3 combines Algorithm 2 with Algorithm 1 to enhance data-adaptive capability. Notice that besides adding a piece of codes of parameter processing, there only make little change to current codes to upgrade the previous framework with our new approach. This is very convenient. Through comparison between step 2 in Algorithm 2 and step 2~3 in Algorithm 3, we can clearly see how our parameter processing algorithm works in framework.

```
Algorithm 2: SSC-OMP
─────────────────────────────────────────────────────────
Input: matrix $X = [x_1, \cdots, x_N]$, $K$, $\varepsilon$.
1: Normalize data X to a range.                    // usually 0~1
2: Compute $c_i^*$ from OMP($X_{-i}, x_i, K, \varepsilon$) using the OMP Algorithm. [21]
3: Set $C^* = [c_1^*, \cdots, c_N^*]$ and $A = |C^*| + |C^{*T}|$.
4: Compute segmentation from $A$ by spectral clustering.
Output: segmentation of data $X$.
─────────────────────────────────────────────────────────

Algorithm 3: Parameter-K Processing + SSC-OMP (data-adaptive SSC-OMP)
─────────────────────────────────────────────────────────
Input: matrix $X = [x_1, \cdots, x_N]$, $K$, $\varepsilon$.
1: Normalize data X to a range.                    // usually 0~1
2: Compute $K^*$ using Algorithm 1 with $X$ and $K$.
3: Compute $c_i^*$ from OMP($X_{-i}, x_i, K^*(i), \varepsilon$) using the OMP Algorithm.
4: Set $C^* = [c_1^*, \cdots, c_N^*]$ and $A = |C^*| + |C^{*T}|$.
5: Compute segmentation from $A$ by spectral clustering.
Output: segmentation of data $X$.
─────────────────────────────────────────────────────────
```

There are no loops but some simple functions in Algorithm 1, such as matrix product, digit rounding, etc. Considering the mean value of output in Algorithm 1 is about the original $K$, it barely brings additional computation and preserves time efficiency. In fact, the time complexity of Algorithm 1 is $O(N^2 \text{dim})$ at most, much lower than the time complexity of Algorithm 2 which is $O(KN^2 \text{dim})$, where dim denotes the data dimension, then the time complexity of Algorithm 3 is also $O(KN^2 \text{dim})$.

In summary, Algorithm 1 can efficiently coordinate with OMP-based SSC and improve the clustering performance, also keep the time efficiency.

## 4    Current & New evaluation measures

In this section, we introduce some widely used evaluation metrics of SSC problems, and defines a new metric Self-Expressive-Affinity (SEA) ratio to measure the sparse representation conversion efficiency from self-expressive matrix to affinity matrix for spectral clustering.

**Clustering accuracy (ACCR%)** The percentage of correctly labeled data points. When number of clusters decreases, ACCR increases.

**Running time (TIME)** Run all clustering tasks using Matlab. When number of clusters, samples per cluster or dictionary size increases, TIME increases.

**Connectivity (CONN)** The CONN denotes whether the data points in each cluster form a connected component of the graph. For an undirected graph with degree matrix D and adjacency matrix W, we use the second smallest eigenvalue of the normalized Laplacian matrix $L = I - D^{-\frac{1}{2}} W D^{-\frac{1}{2}}$ to measure the connectivity of the graph. Theoretically, CONN $\in [0, \frac{n-1}{n}]$, CONN = 0 if and only if the graph is not connected [22]. In our case, we compute the algebraic connectivity for each cluster and take the min value among them as the measure of connectivity. When number of clusters decreases, CONN increases.

**Percentage of subspace-preserving representations (PERC%)** PERC is a direct measure of whether the solution is subspace preserving or not, stands for the percentage of points whose representations are subspace-preserving. Ideally, PERC=100, only if a subspace-preserving solution is given. When number of clusters decreases, PERC increases.

**Subspace-preserving representation error (SSR%)** SSR measures how close the coefficients are from being subspace preserving. A subspace-preserving coefficient matrix gives SSR=0. When number of clusters increases, SSR increases.

We use ACCR and TIME to evaluate the performance of subspace clustering methods, use CONN to evaluate the connectivity of undirected graph, and use PERC and SSR to evaluate the representation quality. Greater ACCR, CONN, PERC or smaller TIME, SSR gives better clustering results.

Next we develop a new evaluation metric to measure the conversion efficiency from self-expressive matrix to affinity matrix.

**Definition** Let G be a graph, denotes the self-expressive matrix of G by C, then use $A = |C| + |C^T|$ to denote the affinity matrix. Finally, the Self-Expressive-Affinity (SEA) ratio, is defined as $\frac{nnz(A)}{2 \times nnz(C)}$, where $nnz(C)$ represents the number of non-zero elements in matrix C.

**Corollary** SEA $\in [0.5, 1]$. SEA = 0.5 if and only if C is symmetric, SEA = 1 if and only if $\forall i \neq j, C_{ij} \wedge C_{ji} \neq 1$, where $C_{ij}$ represents the *(i,j)*th element of matrix C, and $\wedge$ stands for logical conjunction AND.

*Proof.* Because $A = |C| + |C^T|$, hence $nnz(A) \geq nnz(C)$, by definition SEA $\geq 0.5$. SEA = 0.5 if and only if $nnz(A) = nnz(C)$, then C must be symmetric. $A = |C| + |C^T|$ mean $nnz(A) \leq nnz(C) + nnz(C^T) = 2 \times nnz(C)$, by definition SEA $\leq 1$. SEA = 1 if and only if $\forall i \neq j$, if $C_{ij} \neq 0$, then $C_{ji} = 0$. Considering in some cases $C_{ij} = C_{ji} = 0$, it is the equivalent of $\forall i \neq j, C_{ij} \wedge C_{ji} \neq 1$.

To avoid $C_{ij} \wedge C_{ji} = 1$, somehow alleviate redundant representation or wastes of representation chances for affinity matrix A, which means theoretically greater SEA gives better clustering results, thus ideally SEA=1.

# 5  Experiments and performance evaluation

In this section, we describe the datasets that we used for experiments and give details about the experimental settings, then evaluate performances on real-world datasets and verify our preceding theoretical analysis, show that our new approach and algorithm achieves better performances in terms of clustering accuracy (ACCR), running time (TIME), percentage of subspace-preserving representations (PERC), subspace-preserving representation error (SSR), connectivity (CONN), and Self-Expressive-Affinity (SEA) ratio compared with other OMP-based sparse subspace clustering algorithms. Our new approach also performs well in the anti-noise experiments. Experimental hardware environment: Intel Core i7-6700 CPU@3.40 GHz with Intel HD Graphics 530, and Windows 10.0.17134 operating system. Software development platform: Matlab R2017b. (The code and data are available at: https://github.com/zwapimeow/neurips2019) framework and our approach.

## 5.1  Experiments on Extended Yale B Dataset

Extended Yale B dataset [23] contains frontal face images of 38 individuals, each individual has about 64 different photos, taken from the same viewpoint under varying illumination conditions, each of size $192 \times 168$. In experiments, all images are downsampled to $48 \times 42$ pixels and reshaped as vectors of size dim=2016. We randomly pick number of clusters: n $\in$ {5,15,25,35} individuals and take all the images as the data to be clustered, then compute 20 times for each n. Meanwhile we set threshold value: thr = $10^{-6}$ and dictionary size: K = 8. Table 1 shows the evaluation performance of five methods on the face clustering problem: Sparse Subspace Clustering (convex optimization), Orthogonal Matching Pursuit (OMP), Rotated Orthogonal Matching Pursuit (ROMP) [24], our new algorithm+OMP (ours+OMP) and our new algorithm+ROMP (ours+ROMP). The evaluation metrics are: ACCR, TIME, PERC, SSR, CONN and SEA ratio.

In terms of ACCR and SEA, we can observe better performance when adding our algorithm in front of other methods in all cases, and it's always much better than the performance of SSC. Though SSC has best performance of PERC and SSR, our method brings improvement and multiply PERC, and reduce SSR of OMP and ROMP, especially when number of clusters gets greater. Although adding our algorithm increases the amount of computation, but it's barely influence computational efficiency, when number of clusters gets greater, we can even observe the shortening of TIME. Since our algorithm specifically dispises outliers in data, resulting in slightly reduce CONN of subspaces, but basically it's still higher than SSC.

In addition, we also carried out an anti-noise experiment studies ACCR. Set n=5 and K=8, using the imnoise function in Matlab to add Gaussian white noise of mean and variance default to 0 and 0.01 respectively to all images, while the noise rate: sigma ∈ {0, 0.1, ..., 0.9}. The results are plotted in Figure 4, when raising sigma, the ACCR of ours+ROMP always remains 1%~5% higher than ROMP, and when sigma<0.5 the ACCR of ours+OMP always remains 1%~6% higher than OMP. These results are pretty good, because when sigma≥0.6, face images have low quality, generally such images wouldn't be used in clustering problem.

Table 1: Experimental results on Extended Yale B dataset

|  | ACCR (%) | | | | TIME (sec.) | | | |
|---|---|---|---|---|---|---|---|---|
|  | 5 | 15 | 25 | 35 | 5 | 15 | 25 | 35 |
| SSC | 73.98 | 55.42 | 42.99 | 38.30 | 6.80 | 52.27 | 123.57 | 207.08 |
| OMP | 89.13 | 80.35 | 76.70 | 75.23 | 0.34 | **4.57** | 17.00 | 29.68 |
| ours+OMP | **95.59** | 82.16 | 78.99 | 77.44 | **0.32** | 4.68 | **14.41** | **29.14** |
| ROMP | 94.91 | 86.27 | 84.22 | 79.58 | **1.84** | 19.80 | 44.95 | 108.56 |
| ours+ROMP | **95.86** | **89.91** | **84.90** | **80.38** | 1.88 | **16.44** | **44.88** | **87.52** |

|  | SEA (%) | | | | CONN ($\times 10^{-2}$) | | | |
|---|---|---|---|---|---|---|---|---|
|  | 5 | 15 | 25 | 35 | 5 | 15 | 25 | 35 |
| SSC | 61.62 | 64.53 | 63.90 | 63.59 | 3.88 | 2.14 | 0.40 | 0 |
| OMP | 89.27 | 91.22 | 92.09 | 92.59 | 7.94 | 3.75 | 1.09 | 0.08 |
| ours+OMP | **90.63** | **92.09** | **92.77** | **93.14** | **7.81** | **2.59** | **0** | **0** |
| ROMP | 90.68 | 91.94 | 92.60 | 93.07 | 13.14 | 5.65 | 1.22 | 0.24 |
| ours+ROMP | **92.34** | **93.34** | **93.97** | **94.39** | **11.47** | **2.84** | **0** | **0** |

|  | PERC (%) | | | | SSR (%) | | | |
|---|---|---|---|---|---|---|---|---|
|  | 5 | 15 | 25 | 35 | 5 | 15 | 25 | 35 |
| SSC | 26.90 | 0.96 | 1.61 | 0.70 | 7.67 | 17.44 | 27.42 | 34.42 |
| OMP | 2.05 | 0.28 | 0.07 | 0.05 | 12.80 | 18.15 | 20.94 | 22.60 |
| ours+OMP | **2.98** | **0.55** | **0.31** | **0.22** | **11.26** | **17.12** | **19.94** | **21.73** |
| ROMP | 1.78 | 0.14 | 0.03 | 0.01 | 13.04 | 20.01 | 23.54 | 25.76 |
| ours+ROMP | **3.32** | **0.62** | **0.37** | **0.20** | **12.51** | **19.76** | **23.14** | **25.46** |

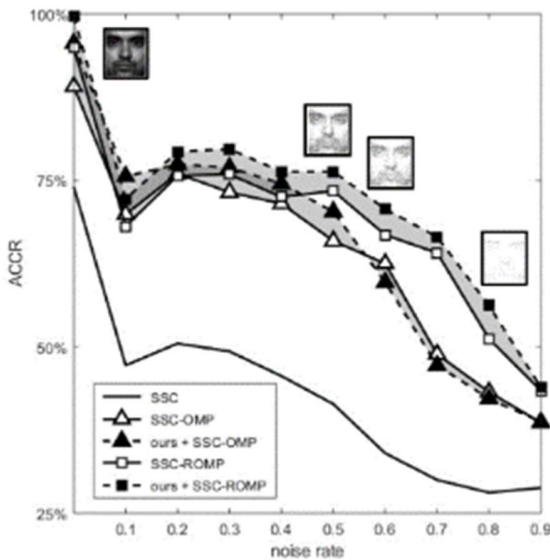

Figure 4: Anti-noise experiments on Extended Yale B dataset

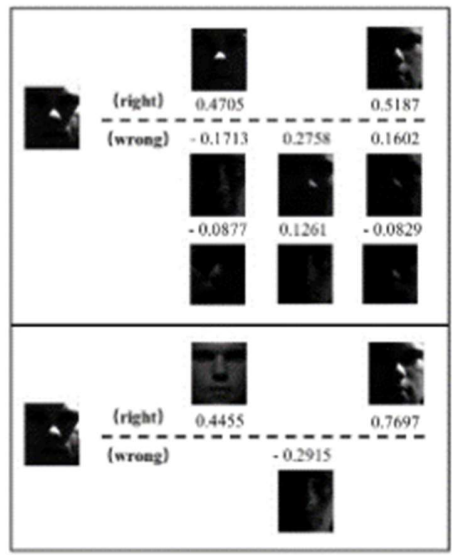

Figure 5: Comparison of OMP (top) and ours+OMP (bottom) on Extended Yale B dataset

Table 2: experimental results on USPS dataset

|  | (sample/cluster) | ACCR (%) OMP | ACCR (%) ours+OMP | TIME (sec.) OMP | TIME (sec.) ours+OMP | SEA (%) OMP | SEA (%) ours+OMP |
|---|---|---|---|---|---|---|---|
| K=8 | 250 | 58.39 | **66.40** | 0.59 | **0.56** | 92.16 | **92.69** |
|  | 500 | 57.48 | **65.68** | 2.28 | 2.33 | 93.70 | **94.08** |
|  | 800 | 51.06 | **60.03** | 9.54 | 10.27 | 94.55 | **94.81** |
|  | 1100 | 48.82 | **55.70** | 27.60 | **24.65** | 94.97 | **95.20** |
| K=16 | 250 | 56.18 | **67.06** | 1.21 | **1.17** | 92.16 | **94.24** |
|  | 500 | 57.14 | **65.72** | 4.29 | **4.25** | 95.29 | **95.73** |
|  | 800 | 49.96 | **59.78** | **18.48** | 18.77 | 96.14 | **96.47** |
|  | 1100 | 43.70 | **53.18** | **42.02** | 44.65 | 96.58 | **96.85** |
| K=24 | 250 | 57.70 | **66.80** | 2.05 | **2.00** | 94.05 | **94.75** |
|  | 500 | 57.05 | **65.73** | 7.01 | **4.47** | 95.87 | **95.88** |
|  | 800 | 47.72 | **57.92** | 27.60 | **27.45** | 96.73 | **97.07** |
|  | 1100 | 42.15 | **50.97** | 65.10 | **64.05** | 97.18 | **97.45** |
| K=32 | 250 | 58.16 | **65.42** | **3.47** | 3.52 | 94.22 | **94.99** |
|  | 500 | 54.47 | **66.00** | 9.41 | **9.39** | 96.16 | **96.65** |
|  | 800 | 47.09 | **58.44** | **34.80** | 38.82 | 97.04 | **97.40** |
|  | 1100 | 42.34 | **50.93** | 115.33 | **90.10** | 97.50 | **97.78** |

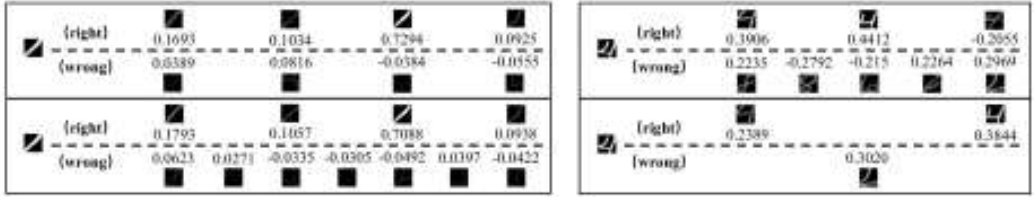

Figure 6: Comparisons of OMP(top) and ours+OMP(bottom) on USPS dataset

### 5.2 Experiments on USPS Dataset

We use a standard dataset for handwritten digit recognition, USPS, consists of 8-bit grayscale images of number 0 through 9, so here we have n=10. All images in USPS dataset are $16 \times 16$ grayscale pixels, we reshape each sample as vectors of size $dim = 256$ and randomly pick sample $\in \{250, 500, 800, 1100\}$ in each cluster, meanwhile set $K \in \{8, 16, 24, 32\}$ and $thr = 10^{-6}$. When we randomly select 5 numbers for testing and take the averages after 20 times of calculation, we get results in Table 2. These results show the difference between Orthogonal Matching Pursuit (OMP) and our new algorithm+OMP (ours+OMP). The evaluation metrics are ACCR, TIME and SEA ratio.

We can observe that ours+OMP always performs better in terms of ACCR and SEA. When K or sample per cluster changes, the ACCR of ours+OMP always remain 7%~12% higher than OMP, and SEA also remain higher than OMP. That is to say when K or sample per cluster changes, the advantages of our approach are unaffected. We also observe that TIME is barely changed, proves that our approach keeps computational efficiency.

### Conclusions

In the paper we propose a novel approach to enhance the data-adaptive capability for OMP-based sparse subspace clustering. A parameter selection process is proposed to expand the existing framework, and a specific parameter processing algorithm is proposed to adjust parameters based on data distribution. Experiments verifies effectiveness, efficiency and anti-noise ability of our approach.